# Interlingual Lexical Organisation for Multilingual Lexical Databases in NADIA

Gilles Sérasset

Gilles.Serasset@imag.fr

GETA, IMAG-campus (UJF & CNRS)
BP 53, 38041 Grenoble Cedex 9, France

**Abstract**

We propose a lexical organisation for multilingual lexical databases (MLDB). This organisation is based on acceptions (word-senses). We detail this lexical organisation and show a mock-up built to experiment with it. We also present our current work in defining and prototyping a specialised system for the management of acception-based MLDB.

## Introduction

Needs for large scale lexical resources for Natural Language Processing (NLP) in general and for Machine Translation (MT) in particular, increase every day. These resources are considered to represent the most expensive part of almost any NLP system. Hence, an increasing interest in the development of reusable dictionaries can be observed.

To develop a Multilingual Lexical Database (MLDB), we think of two main approaches. First, the *transfer approach* where the links between the languages are realized via unidirectional bilingual dictionaries. This approach is used by many MT systems and by some lexical database projects (notably Acquilex or Multilex). Second, the *interlingual approach* where the links between the languages are realized via an unique interlingual dictionary. The KBMT-89 project (Knowledge Based Machine Translation) at Carnegie Mellon University in the US and the EDR (Electronic Dictionary Research) project in Japan use this approach.

In the context of multilingual MT systems, we are interested in the problems posed when constructing and using an application and theory independent MLDB. We are developing a Lexical Database Management System, NADIA, based on an interlingual approach. We chose *acceptions* as interlingual units. NADIA provides many tools for the management of MLDBs. Moreover, this system gives the linguist a great liberty in the choice of the linguistic structures.

We first give an overview of the project, beginning with its lexical organization. Then, we give the results of our experimentations on this lexical organization. Finaly, we present our current work: the definition and prototyping of a specialized system for the management of acception-based MLDBs.

NADIA is the continuation of a work done for the Multilex ESPRIT project. The coherence checker and software architecture have been defined for Multilex and adapted to our lexical organization.

## I. Acception-based lexical organization

After studying and comparing different projects of lexical databases, including EDR (EDR 1993), KBMT-89 (Nirenburg and Defrise 1990; Goodman and Nirenburg 1991) Multilex and of Multilingual MT systems, such as CICC (Uchida and Zhu 1991) and ULTRA (Farwell, Guthrie et al. 1992), we have concluded in favor of an interlingual lexical organization for our MLDBs.

Some of the international projects of lexical databases are based on a multi-bilingual approach (e.g. Multilex) while others use knowledge representation as an interlingua (e.g. KBMT-89 or EDR). Much like ULTRA, our approach is interlingual and linguistic rather than knowledge-based.

### 1. The dictionaries

A MLDB consists of two kinds of dictionaries: the monolingual dictionaries and the acception dictionary.

#### 1.1. Monolingual dictionaries

The *monolingual dictionaries* are accessible by *entries*. These entries are lemmas ("normal form" of words, e.g. in English, infinitive for verbs, singular for nouns, etc.).

Items of the monolingual dictionaries (*monolingual acceptions*) are generally accepted meanings of words or expressions, as we can find them in standard printed dictionaries. These monolingual acceptions are combined with their linguistic information.

Monolingual acceptions of a language L are acceptions that are connected to a word or an expression of L. Such an acception can be accessed from one (or more) entries.

#### 1.2. Acception dictionary

The interlingual dictionary, called *acception dictionary*, contains *interlingual acceptions*. Some information can be linked to these interlingual acceptions.

In a MLDB composed of $n$ monolingual dictionaries, the set of interlingual acceptions is equal to the union of the sets of monolingual acceptions of the $n$ dictionaries, with an equality relation bound to the semantic identity.

Some contrastive problems may appear when two monolingual acceptions of two different languages are semantically slightly different. This appears when there is a non-direct translation of a word (e.g. 'river' can be translated in French by 'rivière' or by 'fleuve'[1]). This kind of problem is solved by a relation from acception to sub-acception which is pre-defined in all NADIA lexical databases: the contrastive relation. It is intended to code contrastive problems induced by a non-direct translation, it is not intended to code any kind of ontological information.

---

[1] A 'rivière' is a rather small river flowing into another river. A 'fleuve' is a large river flowing into the see.

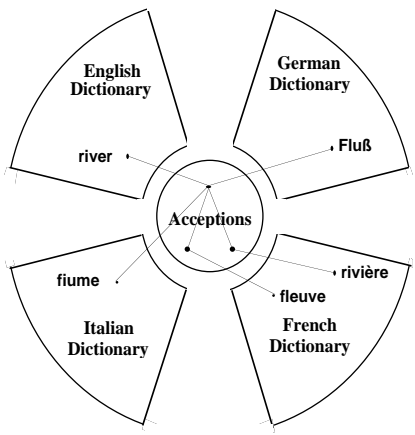

Fig. 1: illustration of the acception-based lexical organization

**2. Lexical organization**

In the acception-based lexical organization, the monolingual acceptions and the interlingual acceptions must satisfy the following criteria:

<u>2.1. Well-formedness criteria</u>

- *Each interlingual acception corresponds to at least one monolingual acception*. This criterion states that an interlingual acception must correspond to at least one entry of one language (as monolingual acceptions).
- *An interlingual acception corresponds to at most one monolingual acception of the same language*. An interlingual acception is not necessarily connected to a monolingual acception of each language of the MLDB.
- *A monolingual acception corresponds to one and only one interlingual acception*. A monolingual acception is always related to an interlingual acception and (as stated by the preceding criterion) is one-one.

<u>2.2. Translation criteria</u>

- *Two monolingual acceptions of different languages correspond to a unique interlingual acception if, and only if, they have the same meaning*. This criterion states the semantic identity of two monolingual acceptions of different languages (provided that they correspond to the same interlingual acception) allowing the use of the interlingual dictionary for lexical translation purposes.

- If entry e1 of language L1 is translated by entry e2 in language L2 via a non-direct equivalence, the corresponding interlingual acception must be linked by the contrastive relation or by a relation of quasi-synonymy. This criterion allows the use of the acception dictionary for lexical translation purposes even when there is no direct translation.

**II. Experimentation**

**1. The Parax mock-up**

In order to experiment this lexical organization, É. Blanc has built the Parax mock-up (Sérasset and Blanc 1993). This mock-up is a small acception-based lexical database of 5 languages (English, French, German, Russian, Chinese).

Parax, produced on Macintosh with HyperCard™, was designed to experiment problems inherent to the acception-based lexical organization. Hence, items of the monolingual dictionaries are combined with rather simple linguistic information.

An entry of a monolingual dictionary is linked to several acceptions. These acceptions are provided with their linguistic information (left column in fig. 2). Each of these monolingual acceptions is related to an interlingual acception along with its definition (in French) and some semantic information (right column in fig. 2).

Fig. 2: Monolingual entry: "épouser" (to marry, to fit, to espouse)

To accede to the acception dictionary, the user selects an acception in the middle column. The acception is displayed along with its sub-acceptions (middle column of fig. 3). From this point, it is possible to get a translation by selecting a target language for one of the acceptions. The translation appears in the right column of fig. 3 (which shows the German translation of the acception). In the example given, there is no direct equivalence from French to Russian as Russian introduces a distinction on the gender of the subject. To get the Russian translation, we have to select one of the sub-acceptions. Then, we can get 'жениться' for a man or 'замуж', 'замуж (выйти - за)' or 'замужем' for a woman.

Fig. 3: Acception: #épouser_semarier (to marry) and it's sub-acceptions.

## 2. Indexing methodology

### 2.1. Indexing in Parax

As the platform we used for this mock-up was not specialized for such a task, we have used an indexing methodology for the construction of this MLDB.

The starting point of our work was a small French corpus we wanted to index. Hence, we began to index French words and for each created acceptions, we gave a translation in the other languages.

After creating an entry, the lexicographer gives its different word-senses and their linked linguistic information (the kind of information depends on the language of the entry).

Then, the lexicographer links the word senses to an interlingual acception. As the number of acceptions is still small, it is possible to select an already existing acception by browsing directly in the acception dictionary. If the searched acception does not exist, it is created along with a definition in French and some semantic information.

### 2.2. General case

When developing a large scale MLDB, it is no longer possible to select existing interlingual acceptions by directly browsing through the acception dictionary. Moreover, the different dictionaries will have to be indexed by different lexicographers. Hence, it is necessary to define another methodology.

The process of creation of an entry and its monolingual acceptions does not change. After creating an entry, the lexicographer selects a possible translation for the considered acception in a language of the database. If this translation is already indexed in the target language, he selects the corresponding acception in the target dictionary. The source and target monolingual acceptions are automatically linked to the same interlingual acception. If the translation is not already indexed in the target language, the lexicographer indexes it (partially) and asks the person in charge of the target dictionary to complete the new entry.

The acception dictionary is thus constructed and managed by the system and the lexicographers work in more or less the same way as when indexing bilingual dictionaries. This automatic management of the interlingual dictionary involves the automatic verification of the criteria defined above.

When a problem is detected the system attaches a warning for the lexicographer in charge of the acception dictionary, and proposes a default solution.

### 3. Some results

The corpus we wanted to index in the Parax mock-up consisted of 135 entries in French corresponding to a representative set of verbs, nouns, adjectives, and adverbs of general vocabulary. We have indexed these entries and the related acceptions. As we started the indexation with a French corpus, only some of the entries in the other languages have been given all their acceptions.

The distribution of the entries and acceptions of the different languages is the following:

|         | Entries | Acceptions |
|---------|---------|------------|
| French  | 135     | 484        |
| English | 304     | 484        |
| German  | 388     | 509        |
| Russian | 394     | 545        |

This represents a total of 589 interlingual acceptions. Among these interlingual acceptions, 58 are sub-acceptions introduced by contrastive problems. The size of this mock-up is of the same order as that of Multilex.

## III. Current work

Our current work consists in the definition and prototyping of a specialized management system for acception based MLDBs.

### 1. Related projects

Some international projects have already started the development of a system for MLDBs. We have studied and we use some of their results.

In Europe, we have participated in the Multilex project (CEC - DG XIII - ESPRIT project) which aims at the definition of standards for lexical databases systems. We use some of its results (e.g. the software architecture, some of the tools).

Multilex's software architecture, based on three layers (presentation level, internal level and database level), clearly separates the presentation from the coding and the coding from the storage of the information. This organization allows to change the presentation of the structures (giving the possibility to define user interfaces hiding the internal structure).

We have also studied the Japanese EDR project which has developed large dictionaries of about 300,000 words in both English and Japanese (200,000 of general vocabulary, 100,000 of terminological vocabulary). EDR has also developed dictionaries of 400,000 concepts, dictionaries of 300,000 co-occurrences (both in English and Japanese) and dictionaries of 300,000 bilingual entries (both for Japanese-English and English-Japanese) (EDR 1993).

In EDR, individual concepts are introduced in the word dictionary and correspond to the word senses. Hence, our acceptions are really close to their concepts. However, they do not use a contrastive relation to code problems between the languages.

The CICC (Center of International Cooperation for Computerization, Japan) has also used a very close organization to construct a MLDB (Japanese, Chinese, Thai, Malay, and Indonesian) for its Multilingual Machine Translation system. This lexical database contains 50,000 words or terms (Uchida and Zhu 1991).

### 2. Toward a specialized management system

A specialized management system for acception-based lexical databases must offer ways to automatise the management of the acception dictionary. It must also offer tools to define, index and manage the monolingual dictionaries.

## 2.1. Overview

The NADIA system has to detect potential errors in the acception database. Each potential error is given to a lexicographer who is in charge of the correction. This detection is independent of the linguistic structure of the monolingual dictionary. It consists in the detection of geometric inconsistencies in the relations between the elements of the database (entries, monolingual acceptions, interlingual acceptions).

The NADIA system also provides tools to help the users define, index, and manage a MLDB.

These tools depend on the linguistic structure of the different dictionaries. Hence, a linguist has to declare the structure of the articles of the dictionaries via a specialized language. To encode the linguistic information, the linguist can use predefined basic data structures (strings, lists, sets, trees, graphs, automata or Typed Feature Structures).

Several tools have been defined to help the users:
- *Editor*: this tool provides a default interface to edit items of a dictionary. It is also possible to customize the interface — this tool is a structured editor à la GRIF (André, Furuta et al. 1989).
- *Browser*: this tool gives ways to browse through the database.
- *Coherence checker*: the linguist may define some coherence and integrity rules that apply on an article, on a dictionary or on the whole lexical database. These rules are checked and the result depends on the strength of the rule.
- *Defaulter*: the linguist may also define rules to default entries of a dictionary. These rules can be applied in batch mode (in order to expand an existing dictionary) or in interactive mode (to help the lexicographer in the indexing process).
- *Import/export*: the linguist may write importing and exporting procedures from the internal structure to an external format based on the SGML language and TEI guidelines.

## 3. Definition of the linguistic structures and coherence checking

As an example of the use of the NADIA management system for acception-based MLDB, we give the definition of the linguistic structure used in the Parax mock-up (see above). Then, we give some constraints that can be defined on this database.

### 3.1. Definition of Linguistic Structure (DLS)

The linguistic structure used in Parax is inspired by the structures of the dictionaries of GETA's ARIANE system. It is a flat list of attribute-value pairs.

#### 3.1.1. An example: Parax "DLS"

We give here a LISP form of the definition of the structures. Other dialects will be defined in order to hide this LISP form to the linguist (see below).

*Definition of the database*

Before defining the structures of a dictionary, the linguist has to define the database. This definition consists of a declaration of the dictionaries contained in it (here, a database called "Parax" with 4 monolingual dictionaries). For each dictionary, the linguist enters its name, its language, its owner, an optional comment and the classes used to code its entries and acceptions.

```
(define-database Parax
  :owner "GETA"
  :comment "This database is the same as the Parax mock-up defined by Etienne Blanc with hypercard."
  :dictionaries
    (define-dictionary French
      :language "Français"
      :owner "GETA"
      :entry 'French-entry
      :acception 'French-acception)
    (define-dictionary English
      :language "English"
      :owner "GETA"
      :entry 'English-entry
      :acception 'English-acception)
    ...)
```

*Structures of the French dictionary*

The linguist defines the linguistic structures of the dictionaries with an object-oriented language. This task is analogous to the definition of classes in an object-oriented language, or to the definition of the structure of a structured document (à la GRIF, LaTeX or FrameMaker).

Two "classes" are already defined by the system: entry and acception. The linguist determines the structures to be associated with these objects. Here, we give the definition of the structure of the French dictionary.

The predefined class *entry* implements a tree with acceptions on its leaves. In the following example, an entry consists in a feature structure with two features (a graphic-form and a category).

```
(def-linguistic-class french-entry (entry)
  (feature-structure
   (graphic-form string)
   (category category)
  ))

(def-linguistic-class category ()
  (one-of 'nc 'np 'vb 'vbimp
          'vbrefl 'adj 'card
          'deict 'repr 'sub 'coord))
```

The predefined class *acception* provides a way to code its relation with an interlingual acception. In the example, we define an acception as a feature structure with features representing derivation information (with the kind and the source of a derivation), information on valencies, etc.

```
(def-linguistic-class french-acception (acception)
  (feature-structure
   (cat category)
   ;; information on the derivation.
   (drvv (feature-structure
          (deriv-kind
           (one-of 'naction 'nresult 'nlieu 'nagent
                   'ninstr 'adject 'adjpass 'adjpotpas
                   'adjresact 'verbe))
          (deriv-from symbol)))
   (drvn (feature-structure
          (deriv-kind
           (one-of 'ncond 'nlieu 'ninstr 'ncollect
                   'nperson 'adjrelat 'adjqual 'verbe))
          (deriv-from symbol)))
   (drva (feature-structure
          (deriv-kind
            (one-of 'nabst 'nperson 'verbe))
          (deriv-from symbol)))
   ;; information on the valencies
   (val0 valency)
   (val1 valency)
   (val2 valency)
   (val3 valency)
   (val4 valency)
   ;; other information
   (gnr (any-of 'masc 'fem))
   (nbr (any-of 'sg 'pl))
   (aux (one-of 'être 'avoir))
   (reciproque (one-of 'arg0-arg1 'arg1-arg1))
   (aspect (one-of 'achevé 'inachevé 'début 'fin
                   'duratif 'fréquent 'instantané))
  ))
```

```
(def-linguistic-class valency ()
  (any-of 'nom 'à+nom 'avec+nom 'comme+nom
    'contre+nom 'dans+nom 'de+nom 'en+nom
    'entre+nom 'par+nom 'parmi+nom 'pour+nom
    'sur+nom 'inf 'à+inf 'de+inf 'adj 'que+ind
    'que+subj 'se-moy 'se-pass 'lieu-stat 'lieu-dyn
    'manière 'zéro))
```

### 3.2. Coherence checking

When the definition of the structure is done, the linguist can define coherence rules that will be applied on the entries.

#### 3.2.1. Three kinds of rule

The linguist can define three kind of rules:

- *Integrity* rules apply to an article of a dictionary. They ensure that none of the article of the lexical database has an ill-formed configuration.
- *Local coherence* rules apply to different articles of the same dictionary. They ensure that the dictionary is coherent.
- *Global coherence* rules apply to different articles of different dictionaries of the lexical database. They ensure some coherence between dictionaries.

#### 3.2.2. Three levels of coherence rules

The rules are associated with a strength:

- *Warning*: when the constraint is overridden, a message is passed to the lexicographer, but all treatments are allowed. The warning disappears as soon as the lexicographer validates the entry. These constraints are used to detect potential errors.
- *Delay*: when the constraint is overridden, the lexicographer receives a message and some treatments are forbidden on the concerned entries. Incorrect entries will not be accessible by extraction requests. Interactive treatments such as browsing and editing are allowed. These constraints are used to handle temporarily incomplete entries.
- *Critical*: these constraints can't be overridden. If a transaction overrides such a constraint, it will be canceled (rollback).

#### 3.2.3. Example of coherence rule declaration

A coherence rule declaration is a method (in the sense of LISP/Common Lisp Object System) which is applied on all objects of the class defined in the parameter list. The body of the rule is a lisp expression that must return T or nil. If the result is nil, the exception mechanism corresponding to the strength of the rule is invoked.

Here is an example of an integrity rule for the French dictionary. This rule verifies that the derivation information is coherent with the category of the acception.

```
(def-integrity drv-cat-coherence
            ((acception french-acception)
             (dictionary french))
            critical
  (cond ((is-one-of (cat acception)
                'vb 'vbimp 'vbrefl)
         (and (empty-p (drvn acception))
              (empty-p (drva acception))))
        ((equal (cat acception)
                'nc)
         (and (empty-p (drvv acception))
              (empty-p (drva acception))))
        ((equal (cat acception)
                'adj)
         (and (empty-p (drvv acception))
              (empty-p (drvn acception))))
        (t t)))
```

### Conclusion

In this paper, we have presented our work on MLDBs. After a study of existing international projects and the definition and testing of the proposed lexical organization, we are currently defining and prototyping a specialized system for the management of acception-based MLDBs: the NADIA system.

This system introduces new interesting points. First, the acception-based lexical organization seems to offer the advantages of an interlingual approach while avoiding some of the theoretical and methodological problems of the knowledge-based approach (Sérasset and Blanc 1993). Second, it gives the linguist the possibility to freely define a collection of linguistic structures with a rather complete set of predefined data structures.

Our objective now is to integrate in this prototype features coming from research in the field of structured documents and a multidialectal facility in all tools, in order to provide lexicographers and other users with an interface in their mother tongue.

### Acknowledgments

A part of this work was conducted in the Multilex project. I wish to thank all Multilex partners and GETA members for their support and feed-back.